\documentclass[letterpaper, 10 pt, conference]{IEEEconf}

\usepackage{graphicx}
\usepackage{bm}
\usepackage{amsmath,amssymb}  % amsthm 和 enumitem 这两个包有冲突，需要注释掉
\usepackage[ruled,linesnumbered]{algorithm2e}
\usepackage{threeparttable}
\usepackage{subfigure} % Choose one: subfigure or subfig
\usepackage{color}
\usepackage{makecell}
\usepackage{multirow}

\usepackage{mathrsfs}
\usepackage{cite}
\usepackage{url}
\usepackage{booktabs}
\usepackage{array}
\usepackage[table]{xcolor}

\usepackage{tikz}
\usepackage{hyperref}  % Hyperlinks should be loaded last
\usetikzlibrary{shapes, arrows, decorations.pathmorphing, patterns}

\graphicspath{{figures/}}

\newcommand{\acc}{\mathbf{a}}
\newcommand{\A}{\mathbf{A}}

\newcommand{\e}{\mathbf{e}}

\newcommand{\F}{\mathbf{F}}

\newcommand{\g}{\mathbf{g}}

\newcommand{\K}{\mathbf{K}}

\newcommand{\M}{\mathbf{M}}
\newcommand{\N}{\mathcal{N}}
\newcommand{\p}{\mathbf{p}}
\newcommand{\R}{\mathbf{R}}
\newcommand{\s}{\mathbf{s}}

\newcommand{\U}{\mathbf{U}}

\newcommand{\x}{\mathbf{x}}

\newcommand{\z}{\mathbf{z}}

\renewcommand{\u}{\mathbf{u}}

\newcommand{\I}{\mathbf{I}}
\newcommand{\zeros}{\mathbf{0}}

\renewcommand{\H}{\mathbf{H}}

\renewcommand{\P}{\mathbf{P}}
\renewcommand{\S}{\mathbf{S}}

\graphicspath{{figures/}}

\newcommand{\mbr}{\mathbb{R}}

\newcommand{\mb}{\mathbf}

\IEEEoverridecommandlockouts

\begin{document}

\title{Vision-Based Cooperative MAV-Capturing-MAV}
\author{Canlun Zheng$^{1,2}$, Yize Mi$^2$, Hanqing Guo$^2$, Huaben Chen$^2$, Shiyu Zhao$^2$
\thanks{$^1$College of Computer Science and Technology, Zhejiang University, Hangzhou, China.}
\thanks{$^2$WINDY Lab, Department of Artificial Intelligence, Westlake University, Hangzhou, China.}
\thanks{\{zhengcanlun, miyize, guohanqing, chenhuaben, zhaoshiyu\} @westlake.edu.cn}
}

\maketitle

\begin{abstract}
MAV-capturing-MAV (MCM) is one of the few effective methods for physically countering misused or malicious MAVs. This paper presents a vision-based cooperative MCM system, where multiple pursuer MAVs equipped with onboard vision systems detect, localize, and pursue a target MAV. To enhance robustness, a distributed state estimation and control framework enables the pursuer MAVs to autonomously coordinate their actions. Pursuer trajectories are optimized using Model Predictive Control (MPC) and executed via a low-level SO(3) controller, ensuring smooth and stable pursuit. Once the capture conditions are satisfied, the pursuer MAVs automatically deploy a flying net to intercept the target. These capture conditions are determined based on the predicted motion of the net. To enable real-time decision-making, we propose a lightweight computational method to approximate the net’s motion, avoiding the prohibitive cost of solving the full net dynamics. The effectiveness of the proposed system is validated through simulations and real-world experiments. In real-world tests, our approach successfully captures a moving target traveling at \(4 \, \mathrm{m/s}\) with an acceleration of \(1 \, \mathrm{m/s^2}\), achieving a success rate of $64.7\%$.
\end{abstract}

% \begin{IEEEkeywords}
%     Cooperative MCM system, vision-based perception, formation pursuing, and automatic launching.
% \end{IEEEkeywords}

\IEEEpeerreviewmaketitle

\section{Introduction}

With the widespread adoption of Micro Aerial Vehicles (MAVs) in surveillance, agriculture, infrastructure inspection, and military applications \cite{sa2017weednet,arias2022middleware}, safety concerns have become increasingly critical \cite{yaacoub2020security}. For instance, MAVs operating near airports or public events pose serious risks, highlighting the consequences of their misuse and the growing need for effective countermeasures.

MAV capture technology spans multiple research areas, including detection \cite{zheng2021air,guo2024global}, state estimation \cite{liu2022vision}, tracking control \cite{li2022three}, and capture strategies \cite{gomez2020sma}. Although significant progress has been made, integrating these modules into a complete interception system remains a challenge. 
Current MAV capture approaches can be broadly classified into signal interference and physical capture methods. Signal interference aims to disrupt control links but suffers from significant limitations, such as crash risks, public safety hazards, and ineffectiveness against autonomous MAVs operating independently of external signals.

\begin{figure}
    \centering
    \includegraphics[width=1\linewidth]{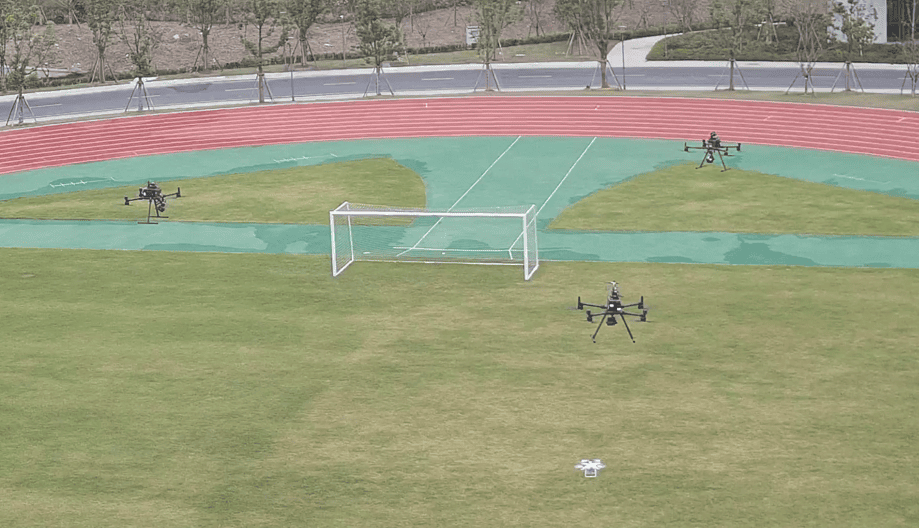}
    \caption{The cooperative MCM system is tracking a moving target MAV. The three black MAVs denote the pursuers. The white MAV denotes the target.}
    \label{fig_capture_system}
\end{figure}

In contrast, physical capture methods, such as nets or robotic claws, offer more direct and reliable solutions. Aerial manipulators equipped with robotic arms can perform precise capture actions \cite{liu2022safely}, but their effective operational range is limited, making them unsuitable for moving targets. Some systems adopt interception through active collision, extending protection coverage but introducing safety risks to the public and the capture MAV itself \cite{gomez2020sma}. On the other hand, net-based capture offers a wider effective range and a faster response time, making it particularly suitable for intercepting fast-moving targets \cite{meng2018net}.
Despite these advancements, developing a fully integrated system capable of real-time perception, tracking, decision-making, and physical capture of dynamic MAV targets remains an open challenge.

In this paper, we propose a cooperative MAV-capturing-MAV (MCM) system, in which multiple pursuer MAVs collaborate to detect, localize, and capture a target MAV, as shown in Fig.~\ref{fig_capture_system}. Compared to single MAV capture strategies, this cooperative approach expands the sensing coverage, increases the capture opportunities, and distributes the computational workload among multiple agents.

Each pursuer MAV is equipped with a visual camera, an onboard computer, and a net-gun device. The decision to launch the net is based on the predicted motion of the flying net. To enable real-time decision-making, we develop an efficient flying-envelope surface model that approximates net motion, avoiding the computational cost of simulating the full flying net dynamics. This allows pursuer MAVs to determine the launch timing in real-time during the pursuit. The proposed system is validated through both extensive simulations and real-world experiments. The experiments demonstrate the system's ability to successfully capture a non-cooperative moving target, providing strong evidence of its effectiveness.

The contribution of this paper is summarized as follows:
1) We develop a vision-based cooperative MCM system that tightly integrates visual perception, state estimation, pursuit control, and autonomous capture decision-making. The collaborative estimation scheme reduces the computational load on individual MAVs while enhancing system robustness.
2) We propose a real-time flying net motion model, enabling pursuer MAVs to predict net behavior and determine capture conditions without requiring full-scale net dynamic simulations.
3) The effectiveness of the proposed cooperative MCM system is thoroughly validated through both simulations and real-world experiments.

This study represents a critical step in enhancing the safety and reliability in capturing non-cooperative moving target scenarios. To the best of our knowledge, this is the first vision-based cooperative MCM system reported in the literature.

\section{Related Work}

\subsection{MAV physical-capture systems}

Net capture and robotic arms are the two most commonly used approaches in MAV physical-capture systems \cite{meng2018net,liu2022safely,yu2023catch}. Compared to robotic arms, tethered nets have gained popularity due to their flexibility, lightweight design, and cost-effectiveness \cite{shan2016review}.

The concept of using nets for capture originates from research on space debris removal \cite{shan2016review}, which has inspired extensive studies on net deployment dynamics. Various modeling approaches, such as the mass-spring model \cite{botta2017deployment, shan2017deployment} and the Absolute Nodal Coordinates Formulation (ANCF) \cite{shan2017deployment}, have been explored to simulate net motion. However, these models are computationally intensive and often require hours to simulate a net with just a few hundred nodes. This high computational cost makes them impractical for MAV interception, where the target is highly agile, and the capture window is fleeting.

To enable real-time capture in dynamic pursuit scenarios, a computationally efficient method is required to predict net motion.

\subsection{Target perception and estimation}

Reliable target detection serves as the foundation for effective pursuit and capture. In recent years, vision-based MAV detection has gained significant attention due to its superior flexibility, robustness, and accuracy over traditional methods such as radar, thermal imaging, and RF detectors \cite{park2021survey}.
Vision-based MAV detection can be classified into appearance-based methods and motion-based methods. Appearance-based methods usually apply state-of-the-art object detectors such as the YOLO series, the R-CNN series, the SSD, and the DETR for aerial target detection \cite{2021DT-Benchmark, zheng2021air, 2022Anti-UAV-DT}. To improve detection accuracy in challenging conditions, some works introduce motion features for aerial target detection, such as optical flow \cite{2021Dogfight, wang2023RAFT}, background subtraction \cite{2021Fast, guo2024global}, and spatial and temporal information \cite{2022Camera, Xie2021SmallLT}.

After detecting the target, accurate state estimation is required. The Kalman filter is a widely used approach for this task, with the pseudo-linear Kalman filter being particularly suited for vision-based sensors. However, ensuring observability often requires continuous high-order maneuvers \cite{li2022three}. To alleviate these maneuvering demands, incorporating bounding box size measurements and the target’s physical dimensions into the state estimation process has been proposed, improving observability conditions \cite{ning2024bearing}.

In contrast, cooperative estimation uses complementary observations from neighboring MAVs, where each agent incorporates not only its own measurements but also information shared by others, thereby improving the accuracy and robustness of state estimation \cite{kamal2015distributed,xiao2021optic,zhuge2023markerless,liu2022vision}. 
Cooperative estimation methods generally fall into two categories: (1) Distributed Kalman filter-based approaches, where state-of-the-art algorithms such as CMKF \cite{olfati2009kalman}, CIKF \cite{olfati2005consensus}, and HCMCI-KF \cite{battistelli2014consensus} improve the accuracy of the estimation by sharing different information and consensus strategies throughout the distributed network.
(2) Distributed Recursive Least Squares (DRLS) methods offer improved computational efficiency and reduced communication overhead. These methods allow for flexible estimator design through customized objective functions, balancing estimation accuracy and communication cost \cite{mateos2012distributed,rastegarnia2019reduced,zheng2023optimal}.

\section{System Overview}

\begin{figure*}
    \centering
    \includegraphics[width=0.9\linewidth]{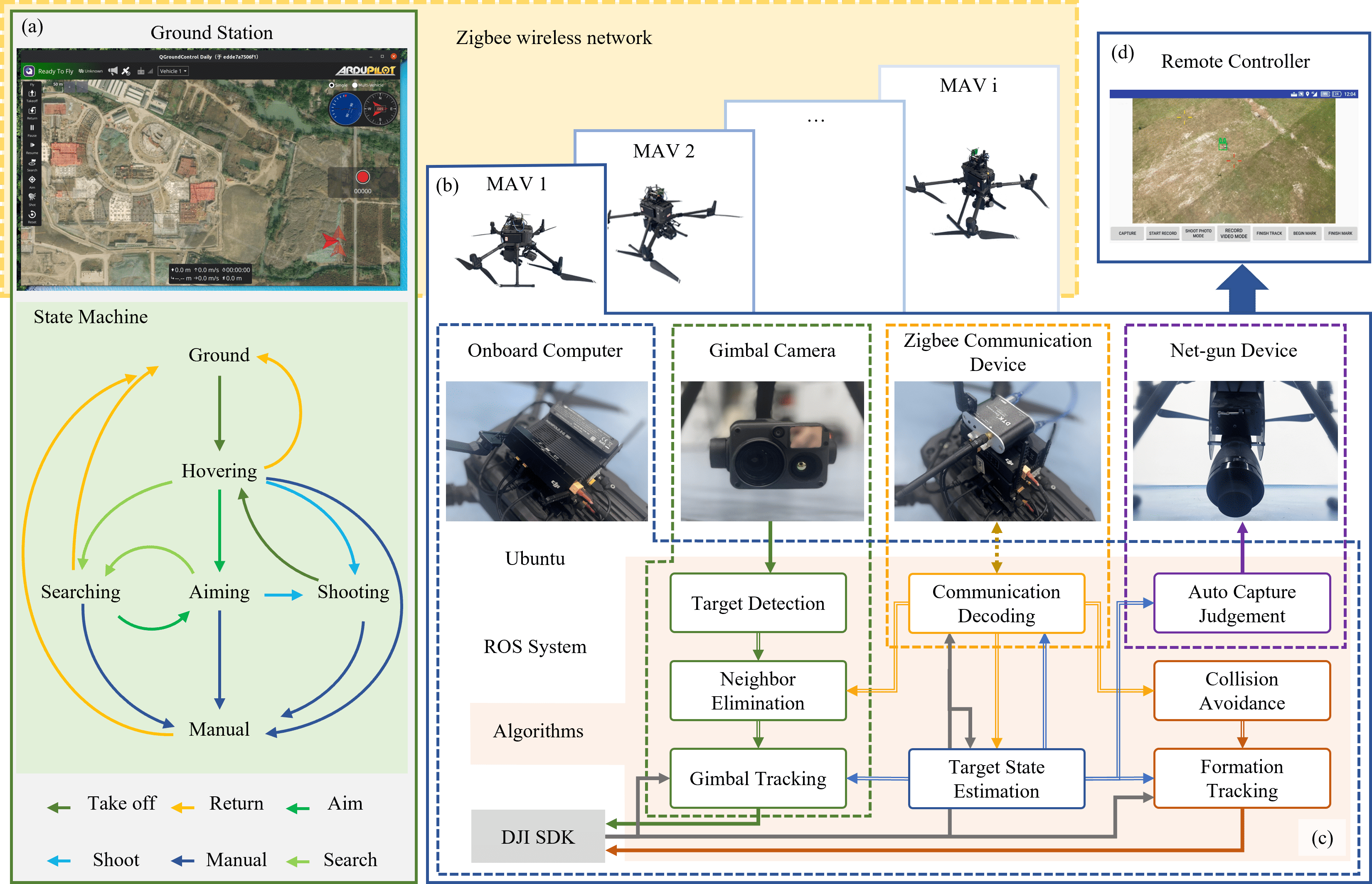}
    \caption{The cooperative MCM system configuration. 
    }
    \label{fig_configuration}
\end{figure*}

The architecture of the proposed cooperative MCM system is depicted in Fig.~\ref{fig_configuration}. This system comprises multiple pursuer MAVs and a ground station, utilizing a distributed network architecture with Zigbee communication devices to facilitate data exchange and command transmission. 

As illustrated in Fig.~\ref{fig_configuration}(a), the ground station interface enables real-time monitoring of both the estimated target and the positions of the pursuer MAVs. The system supports six distinct control modes that can be selected via commands issued from the ground station. Fig.~\ref{fig_configuration}(b) provides an overview of the hardware configuration of each pursuer MAV, which is equipped with an onboard computer, a gimbal camera, a Zigbee communication device, and a net-gun device.  
Computational processes run on an Ubuntu-based onboard computer with the Robot Operating System (ROS) facilitating data exchange, while the Software Development Kit (SDK) enables image stream retrieval and platform control.  
Fig.~\ref{fig_configuration}(c) delineates the communication framework among key algorithmic components. Furthermore, as illustrated in Fig.~\ref{fig_configuration}(d), real-time visual feedback and detection results are accessible via a remote controller.

The capture process follows a structured workflow. The gimbal camera first detects the target, with a neighbor elimination algorithm and a gimbal tracking algorithm ensuring the target remains centered while filtering nearby pursuer MAVs. A collaborative state estimator then fuses detection data with observations from neighboring MAVs, combining spatial and temporal information to improve estimation accuracy. This collaborative approach reduces the need for individual agents to perform aggressive observability maneuvers.

Once the target state is estimated, the pursuer MAVs form a circular formation around the target, maintaining a safe distance and ensuring collision avoidance. MPC optimizes smooth trajectories, minimizing body oscillations that could degrade detection and estimation accuracy. In shooting mode, each MAV performs a real-time net motion prediction, dynamically calculating the net's capturing region. When the target enters this region, the MAV autonomously triggers the net launch to intercept the target.

\section{Target Detection and State Estimation}
In this section, we present \emph{perception} of the pursuer MAV, which involves MAV detection and collaborative state estimation.
Accurate target's state estimation is fundamental for effective tracking and capture, as it directly influences the reliability and precision of the system. This process relies on a perception module that integrates target detection, gimbal tracking, neighbor elimination, and state estimation. Therefore, ensuring both stability and accuracy requires a cohesive approach that carefully considers the interplay among these interconnected components. The proposed approach is based on our previous research in target detection \cite{guo2024global} and cooperative state estimation \cite{zheng2023optimal}. In addition, we propose the neighbor elimination algorithm to avoid mistaken detection for neighbor pursuers.

\subsection{MAV detection}
This project adopts a global-local detection strategy inspired by \cite{guo2024global} for fast and reliable detection of target MAVs. First, a YOLOv5s detector scans the full image to locate the target. Once detected, a localized YOLOv5 operates in a \( 320 \times 320 \) cropped region around the last known position, improving resolution and detection accuracy while reducing computational cost. Continuous tracking is maintained using a gimbal camera, which dynamically adjusts pitch and yaw to keep the target centered. With the roll angle fixed at zero, independent PID controllers track the pitch and yaw angles, ensuring stable target alignment and consistent image acquisition.

\subsection{Neighbor elimination}
When the gimbal cameras move, adjacent MAVs might enter the view, leading to false detection. This occurs because they are mistakenly recognized as targets. Such inaccuracies interfere with the gimbal's tracking control and undermine the precision and stability of state estimation, thereby negatively affecting the system's overall efficiency. To mitigate this issue, we map the positions of cooperative MAVs onto the image and remove potential detection bounding boxes based on the overlap degree. Initially, each pursuer MAV receives the positions of other cooperative MAVs via the communication network. The projection of the $j$th neighboring MAV onto the $i$th camera's coordinate system is computed as follows:
\begin{align*}
    \p^{{\rm cam},i}_{j} = (\R^w_{{\rm cam},i})^{-1}(\p^w_j - \p^w_{{\rm cam},i} +\R^w_{b,i} \Delta \p^b_{\rm cam}),
\end{align*}
where $\p^w_j$ is the neighbor $j$th MAV's position in the world frame, $\p^w_{{\rm cam},i}= \R_{b,i}^w \Delta \p_{\rm cam}^{b} + \p^w_i$ is the $i$th camera position in the world frame. $\Delta \p^b_{\rm cam}$ is the position of the camera in the MAV body frame, \(\R^w_{{\rm cam}, i}\) and \(\R^w_{b, i}\) are the rotation matrices from the $i$th camera frame and the MAV body frame to the world frame, respectively. The superscripts \(\{ \rm w,b, cam \}\) represent the variables in the world, body, and camera coordinate systems, respectively. 
The image plane projection is derived using the camera's intrinsic parameters as follows:
\begin{align*}
    \p^{{\rm pic},i}_{j} = \K^{-1} \g^{{\rm cam},i}_j/(\e_3^T\g^{{\rm cam},i}_j),
\end{align*}
where \(\g^{{\rm cam},i}_j = \p^{{\rm cam},i}_{j}/\|\p^{{\rm cam},i}_{j}\|_2 \) is the bearing vector of the neighbor \(j\), $\e_3 = [0,0,1]^T$. \(\K\) is the intrinsic matrix of the camera.

When the projection intersects with existing bounding boxes beyond a set threshold, it is marked as a false detection linked to a neighboring MAV. The process continually narrows down the options until a distinct, non-overlapping bounding box is found, which is subsequently utilized as the detection outcome for the target in the estimation algorithm.
\begin{figure}
    \centering
    \includegraphics[width=1\linewidth]{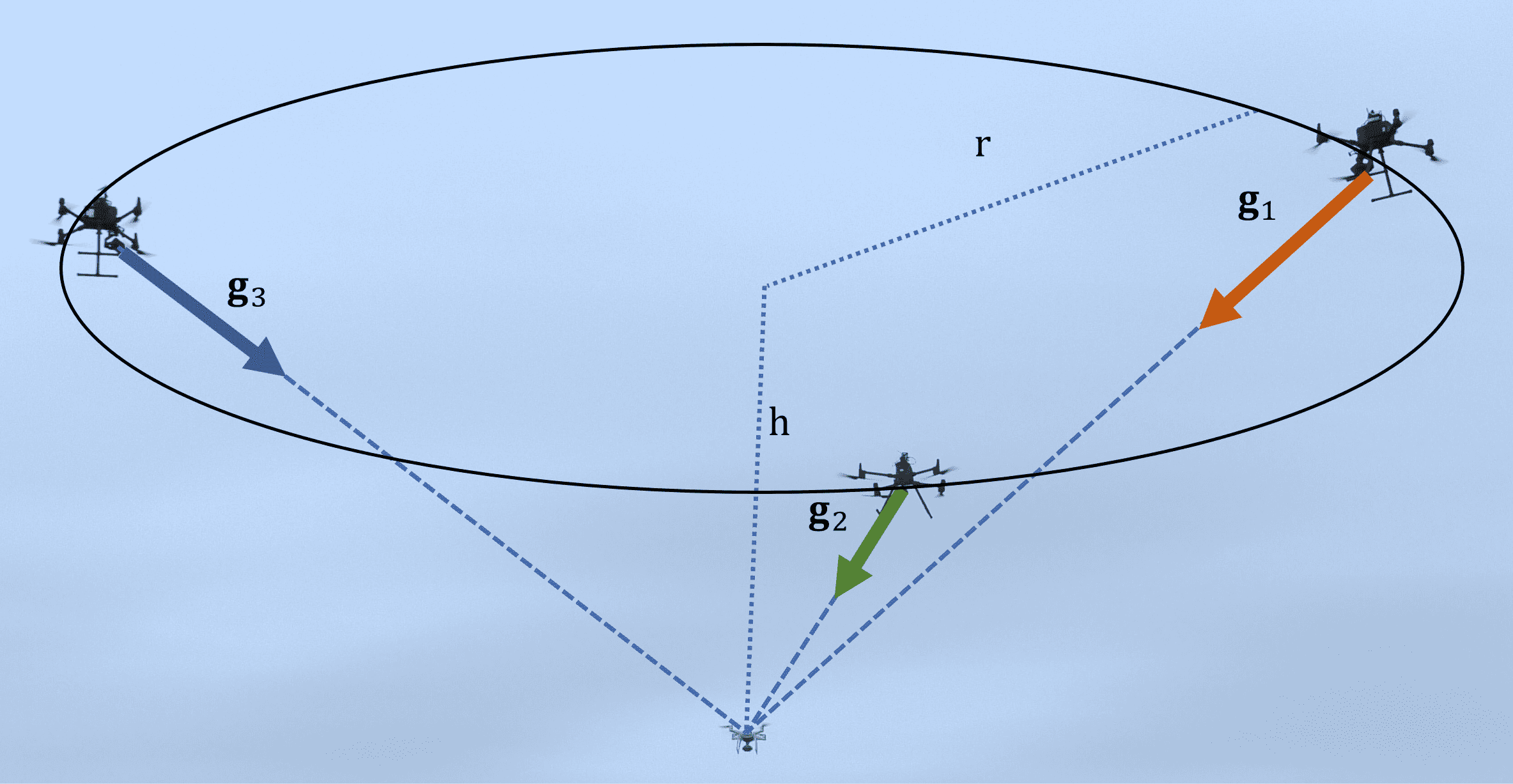}
    \caption{The formation of three cooperative MAVs pursuing a target MAV.}
    \label{fig_formation}
\end{figure}
\subsection{Vision measurements}
The detection bounding box can be converted into a bearing measurement within the world coordinate system using the camera's intrinsic matrix and the orientation of the gimbal camera, as follows:
\begin{align*}
\g^{w}_{{\rm T},i} = \R_{{\rm cam},i}^{w}\K \p_{i}^{\rm cam},
\end{align*}
where $\p_i^{\rm cam} = [x_{{\rm T},i},y_{{\rm T},i},1]^T$ is the target's image position.
We will exclude the superscripts that denote the coordinate system, since all subsequent information is in the same world coordinate system. \( \g_{{\rm T},i} \) represents the bearing measurement of the target by the \( i \)th pursuer MAV. In the following,  \( \g_i \) will be substituted for \( \g_{{\rm T},i} \) for simplicity.
Bearing measurements exhibit significant non-linearity in relation to the target's state, which makes linearization prone to divergence in estimation \cite{li2022three}. Conversely, employing a pseudo-linear approach yields a more reliable estimation performance, as represented by:
\begin{align}
    \underbrace{\P_{\g_i}\s_i}_{\z_i} = \underbrace{\begin{bmatrix}
        \P_{\g_i} & \zeros_3
    \end{bmatrix}}_{\H_i}\x,
\end{align}
where $\x$ is the target's state, including position and velocity. $\z_i$ and $\H_i$ are the pseudo-linear measurement and the measurement matrix, respectively. $\P_{\g_i} = \I_3 - \g_i^T \g_i$ is the projective matrix of $\g_i$. The next step is to estimate the target's state $\x$ with additional neighbor information.

\subsection{Cooperative state estimation}
We use the \emph{Spatial-Temporal Triangulation} (STT) algorithm in the cooperative MCM system, which can make full use of all the available information \cite{zheng2023optimal}.
The STT algorithm is given as follows
\begin{align*}	
	&\hat{\x}^{-}_{i,k}  = \A\hat{\x}_{i,k-1},\\
	&\hat{\M}^{-}_{i,k}  =  \frac{1}{\gamma_1}\left(\A\hat{\M}_{i,k-1}\A^T\right)^{-1},\\
	&\mathbf{e}^{\rm meas}_{i,k}  =  \sum_{j\in \N_i} \H_{j,k}^T\R\left(\z_{j,k}  - \H_{j,k}\hat{\x}_{i,k}^{-} \right),\\
	&\mathbf{e}^{\text{cons}}_{i,k}  = \frac{1}{|\N_i|}\sum_{j\in \N_i}\left(\hat{\x}_{j,k}^{-} -\hat{\x}_{i,k}^{-}\right), \\
	&\S_{i,k}  =c  \sum_{j\in \N_i}\H_{j,k}^T\R\H_{j,k}+ \I_6  ,\\
	&\hat{\M}_{i,k}  = (\gamma_2\hat{\M}_{i,k}^{-}+ \S_{i,k})^{-1},\\
	&\hat{\x}_{i,k}  = \hat{\x}_{i,k}^{-}  + \hat{\M}_{i,k} ( c\mathbf{e}^{\rm meas}_{i,k} + \mathbf{e}^{\text{cons}}_{i,k}).
\end{align*}
Here, $\A$ represents the state transition matrix, while $c$, $\gamma_1$, and $\gamma_2$ are constants with positive values. The terms $\mathbf{e}^{\rm meas}_{i,k}$ and $\mathbf{e}^{\rm cons}_{i,k}$ denote the errors associated with measurement and estimation, respectively. Details of the other variables can be found in \cite{zheng2023optimal}. In executing the STT algorithm, each pursuer MAV acquires \(\{\g_{j,k}, \p_{j,k}, \hat{\x}^{-}_{j,k}\}\) from neighboring agents. 
%By exchanging \(\{\g, \p\}\) instead of \(\{\z, \H\}\), the communication bandwidth is conserved, allowing for an increase in communication frequency and diminished real-world packet loss, as \(\{\z, \H\}\) can be easily reconstructed from \(\{\g, \p\}\).

\section{Formation Tracking and Auto-Capture}
This section presents a formation-based strategy and automatic net capture system. After detecting the target and estimating its state, a surrounding formation is established to guide the pursuer MAVs’ positioning and motion planning, with the estimated target state serving as the reference. The formation control strategy are designed to satisfy three key requirements: 1) Maintaining high observability to ensure accurate target state estimation.
2) Keeping the target within the capture zone long enough to verify the capture conditions.
3) Ensuring stable pursuit control under noisy estimates to preserve formation stability and avoid unnecessary disturbances.

When the target remains in the capture zone for the required duration, the net-launching mechanism is triggered, and the flying net is deployed.

\subsection{Formation design and pursuit control}

To meet the first and second requirement, the pursuer MAVs form a uniform surrounding formation along a horizontal circle, ensuring the bearing-based estimation satisfies observability conditions (see Fig.~\ref{fig_formation}). The three pursuer MAVs continuously adjust their positions to keep the target centered. This positioning ensures the target stays within the capture zones of all pursuers. Additionally, the pursuer MAVs’ yaw angles remain aligned with the target's orientation to ensure the fixed forward-facing net guns are properly aimed.

To meet the third requirement, we developed an MPC-SO(3) pursuing controller. The MPC delivers a reliable anticipated linear acceleration for target pursuing. 
The expected trajectory for the pursuer MAV is then divided into $K$ steps \(\{\x_{{\rm exp}, i}(k)\}_{k=1}^{K}\), where $\x_{{\rm exp},i}(0) = \hat{\x}_i +
[ \Delta \p_i^T, \zeros_{1\times 3}]^T$. $\Delta \p_i$ represents the predefined relative position of the $i$th pursuer MAV with respect to the target MAV.
The objective function of the MPC is expressed as
\begin{align}
    J_i(\U_i)  = \min_{\U_i} \sum_{k=1}^{K}\left(\|\p_{{\rm exp},i}(k) -\right.& \left.   \p_{i}(k) \|^2_\mathbf{Q} +\|\u_{i}(k)\|_{\R}^2 \right) ,\label{eq_J_U}\\
     {\rm subject~to} \quad \x_{{\rm exp},i}(k+1) & = \A\x_{{\rm exp},i}(k),\nonumber\\
    \x_{{\rm MPC},i}(k+1) & =\A \x_{{\rm MPC},i}(k) +\mathbf{B} \u_{i}(k),\nonumber
\end{align}where \(\U_i = \{\u_{i}(t)\}_{t=0}^K\) denotes the control inputs at each individual time step. 

An analytical solution of the control inputs \( \U_i\)  can be derived due to the linearity of the problem \eqref{eq_J_U}, thereby achieving the anticipated acceleration for the pursuer MAV. Given the determination of the desired acceleration and yaw angle, the SO(3) controller is particularly effective for tracking low-level high-frequency commands \cite{lee2010geometric}.

\subsection{Flying net dynamics}

\begin{figure}
    \centering
    \includegraphics[width=0.95\linewidth]{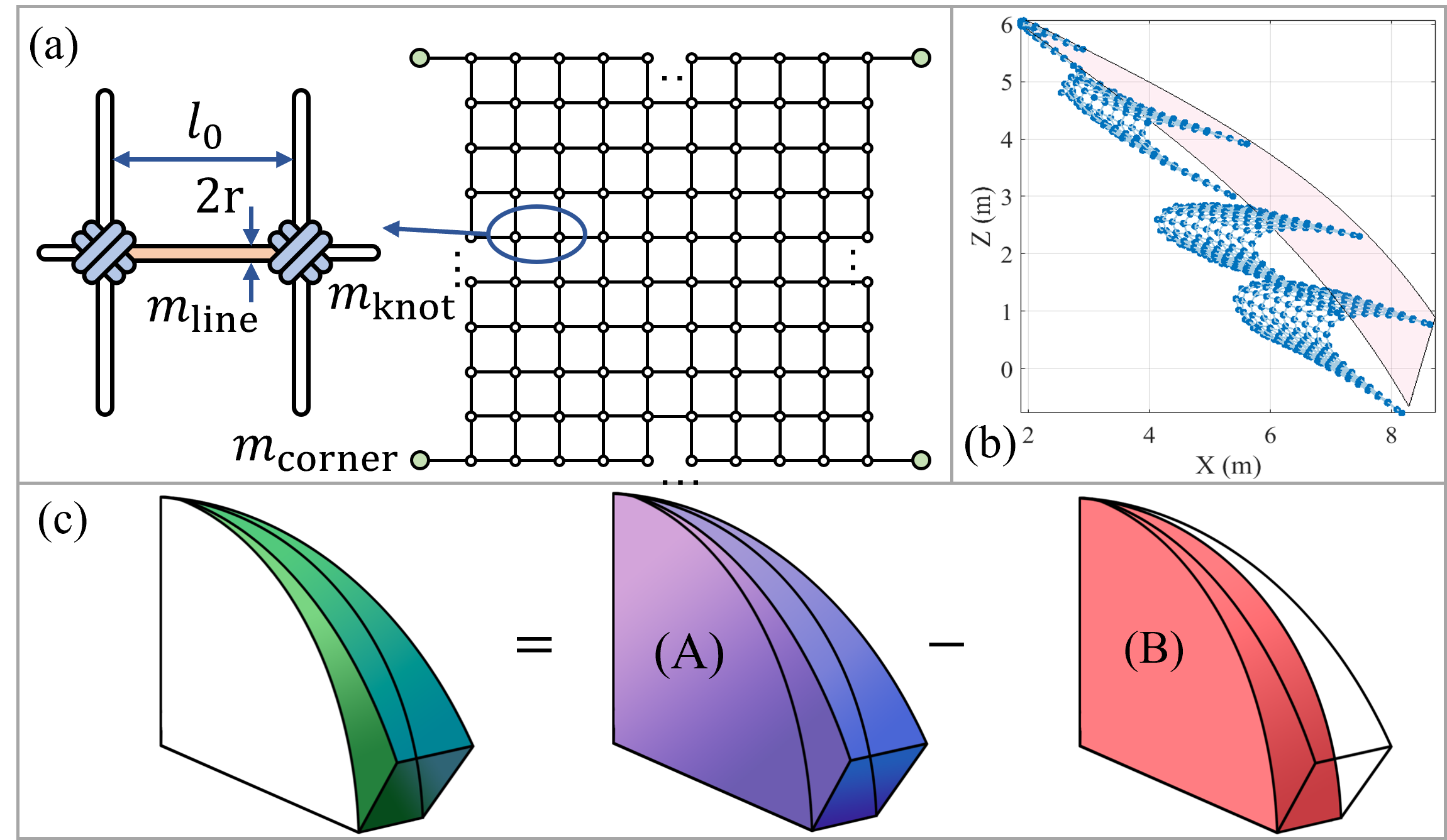}
    \caption{(a) The net structure; (b) The net dynamics capture zone; (c) The capture area of a flying net can be divided into two convex subspaces.}
    \label{fig_net}
\end{figure}

The capture net typically consists of a series of nodes and threads, with several mass blocks attached at its four corners via corner threads, shown in Fig.~\ref{fig_net}(a). Let \( N_s \) denote the number of nodes per side of the capture net. The total nodes, including the corner masses, are \( N = N_s^2 + 4 \). Based on the connections of the nodes in the net, the mass of each node is given as 
\begin{equation}
	m_i =
	\begin{cases}
		|\mathcal{N}_i| m_{\rm line}/2 + m_{\rm knot}, & \text{if } i \in \text{knot nodes} \\
		|\mathcal{N}_i| m_{\rm line}/2 + m_{\rm corner}, & \text{if } i \in \text{corner nodes}
	\end{cases}
	\label{eq:mass_distribution}
\end{equation}
where $\N_i$ denotes the set of threads connected to node $i$. The mass of each thread adjacent to node $i$ is given by $m_{\rm line} = \rho_{\rm net} \pi r^2l_0$, where $\rho_{\rm net}$ represents the density of the net material, $r$ the radius of the thread, and $l_0$ the length without stretch of the thread. $m_{\rm knot} = \rho_{\rm knot} \pi r^2l_{\rm knot}$ is the mass of a physical knot of the net.  A schematic illustration can be found in Fig.~\ref{fig_net}(a). 

The movement of the net suffers from internal tension force, external forces, and gravity. By Newton's second law the dynamic equations of the system for the $i$th node can be formulated as: 
\begin{equation}
m_i\acc_i = \sum_{j\in \N_i}\mathbf{T}_{ij}+ \F_{{\rm ext},i} + \mathbf{G}_i \label{dynamical_equation_1}
	\quad i=1, \ldots, N,
\end{equation}
where $\mathbf{a}_{i}$, $\F_{{\rm ext},i}$, and $\mathbf{G}_i\in\mbr^3$ are the absolute acceleration, the external forces, and the gravity force of the $i$th node, respectively. $\mathbf{T}_{ij}\in\mbr^3$ is each of the tension forces in the node $k$ (belonging to the set $\N_i$) to the node $i$, which can be modeled based on a piecewise Kelvin-Voigt model \cite{christensen2013theory}:
\begin{equation}
	\mathbf{T}_{ij}=\left\{\begin{array}{ll}
		T_{ij} \mathbf{e}_{ij} & \text { if }\left(l_{ij}>l_{0}\right) \wedge\left(T_{ij}>0\right) \\
		0 & \text { if }\left(l_{ij} \leq l_{0}\right) \vee\left(T_{ij} \leq 0\right)
	\end{array}\right. \label{KV model}
\end{equation}${T}_{ij}\triangleq k_{j}\left(l_{ij}-l_{0}\right)+c_{j} v_{ij}^{e}$, with $k_{j}\triangleq\left(E_{\text{net}} \pi r_{j}^{2}\right) / l_{0}$ and $c_{j}\triangleq2 \xi_{a} k_{j} / \omega_{n 1, a}$ are the stiffness and damping coefficients, respectively. $E_{net}$ is the Young's modulus of the net material, $\xi_{a}$ is the chosen damping ratio, and $\omega_{n 1, a}$ is the first natural frequency of the net. $\e_{ij} \in \mbr^3$ and $l_{ij}$ are the direction vector and the relative distance between the $i$th and the $j$th nodes. The projection of the relative velocity $v_{ij}^{e}$ in $\mathbf{e}_{ij}$ can be formulated as:
\begin{align*}
    v_{ij}^{e} & = \lVert\mb{v}_{ij}^{e}\rVert = \lVert\mb{v}^T_{ij}\mb{e}_{ij}\mb{e}_{ij}\rVert,
\end{align*}
where $\mb{v}_{ij} = \mathbf{v}_{j}-\mathbf{v}_{i} \in\mbr^3$ indicate the relative velocity between nodes $i$ and $j$.

The external forces $\F_{{\rm ext},i}$ are primarily the aerodynamic drag exerted by the wind. Therefore, we represent them here as 
\begin{align*}
\F_{{\rm ext},i}=\sum_{j\in \N_i}C_d\lVert\mb{v}_{ij}\rVert^2\frac{\mb{v}_{ij}}{\lVert\mb{v}_{ij}\rVert}= \sum_{j\in \N_i}C_d\lVert\mb{v}_{ij}\rVert\mb{v}_{ij},
\end{align*}
where $C_d$ is the drag coefficient. 
\begin{figure*}
    \centering
    \includegraphics[width=0.9\linewidth]{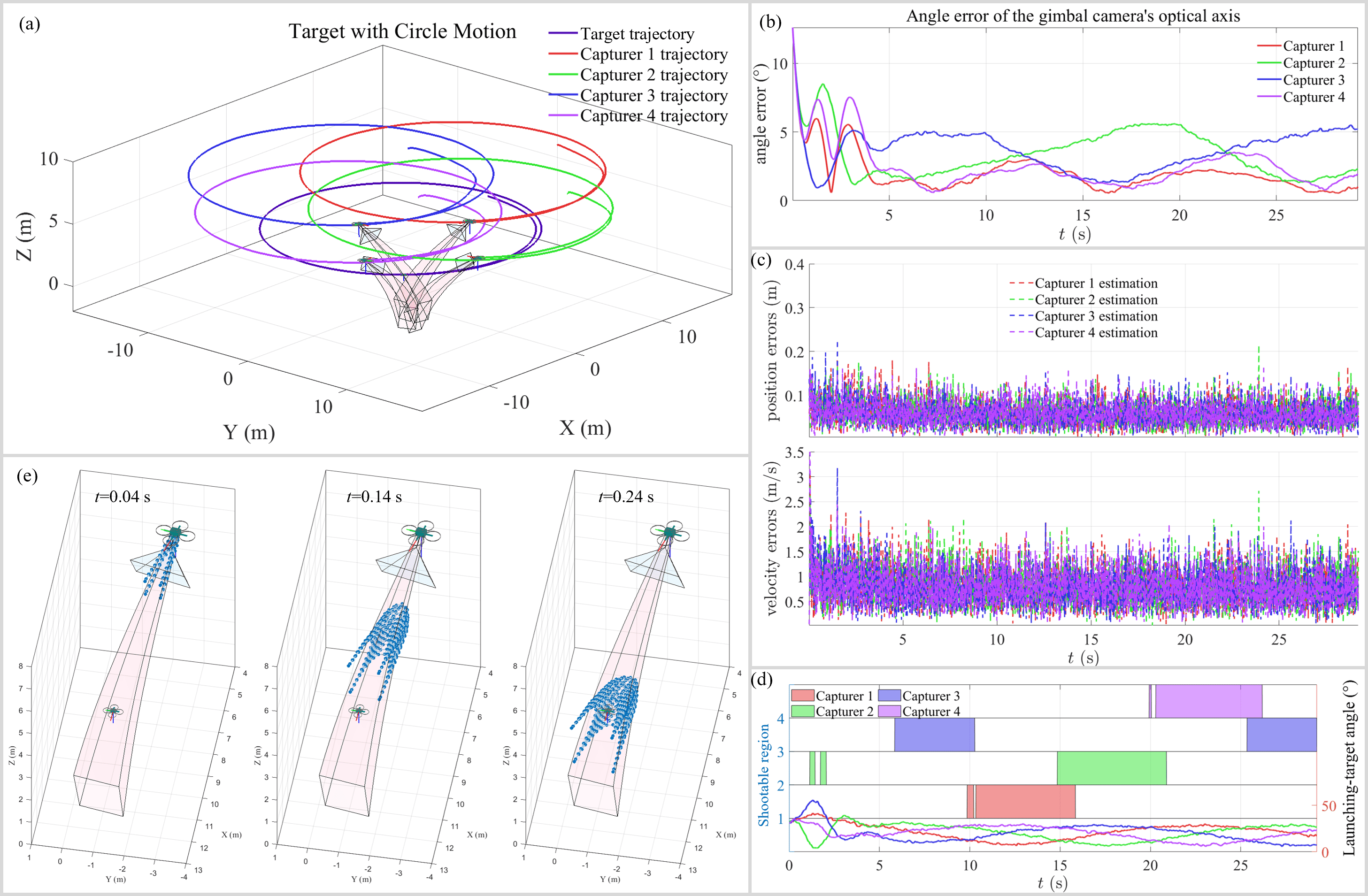}
    \caption{The simulation results of 4 pursuer MAVs cooperative tracking a target with circular motion. 
    }
    \label{fig_sim_results}
\end{figure*}
Equation \eqref{dynamical_equation_1} can be reformulated in a state-space form, allowing the calculation of the position and velocity of each node. Specifically, let $\mathbf{r}=\left[\mathbf{r}_1^T  \ldots  \mathbf{r}_N^T\right]^T\in\mathbb{R}^{3N}$ and $\mathbf{v}=\left[\mathbf{v}_1^T  \ldots  \mathbf{v}_N^T\right]^T\in\mathbb{R}^{3N}$ be the position and velocity vectors of the nodes of the net and the corner masses, respectively.
Define $\mathbf{s}=[\mathbf{r}^T,  \mathbf{v}^T]^T\in\mathbb{R}^{6N}$. Then, the derivative of $\s$ is 
\begin{equation}
	\dot{\mathbf{s}}=\left[\begin{array}{c}
		\mathbf{v} \\
		\mathbf{M}^{-1}\left(\mathbf{T}+\mathbf{F}_{e x t}+\mathbf{G}\right)
	\end{array}\right],\label{dyna_compact}
\end{equation}where $\mathbf{M}\in\mathbb{R}^{3N\times 3N}$ is the mass matrix for all net nodes.

In practice, the net consists of 
$N=365$ nodes, resulting in an extremely high-dimensional system in \eqref{dyna_compact}. Moreover, due to the nonlinearity and discontinuity introduced by the tension forces, solving the full dynamical equation \eqref{dyna_compact} requires substantial computational resources. The high-dimensional nature and complex internal forces make the computation extremely time-consuming, rendering real-time solving impractical.
To overcome this challenge, we propose a simplified net model to approximate the net motion, enabling capture decisions in real-time.
\subsection{Real-time decision-making for net capture}

The key to successful net capture is ensuring the target enters the region defined by the four corner nodes of the net. When a successful capture occurs, the target is always enclosed within this region. In contrast, failed captures typically result from the target escaping this region. Therefore, the capture decision can be made by tracking the motion of only the four corner nodes, rather than all 365 net nodes.

The trajectories of these four corner nodes form a flying envelope surface, illustrated in Fig.~\ref{fig_net}(b). Thus, the pursuer MAV can determine whether to trigger net deployment based solely on whether the target lies within the capture region defined by the four corner nodes, eliminating the need to compute the full net dynamics. Although this approximation may sacrifice some accuracy in modeling the internal interactions between net nodes, it significantly reduces the computational burden, enabling real-time capture decision-making during pursuit.

Furthermore, we decompose the envelope (green area in Fig.~\ref{fig_net}(c)) into two convex subspaces, $\mathbf{A}$ (purple) and $\mathbf{B}$ (red). A target is considered capturable if it is within $\mathbf{A}$ but outside $\mathbf{B}$. This transformation converts the capture decision into a standard point-in-convex-shape test, which is computationally efficient and easy to implement. The capture device is activated only if the target remains in the capture space for at least 0.5~s, mitigating estimation errors and preventing unintended captures. 

\begin{figure*}
    \centering
    \includegraphics[width=0.9\linewidth]{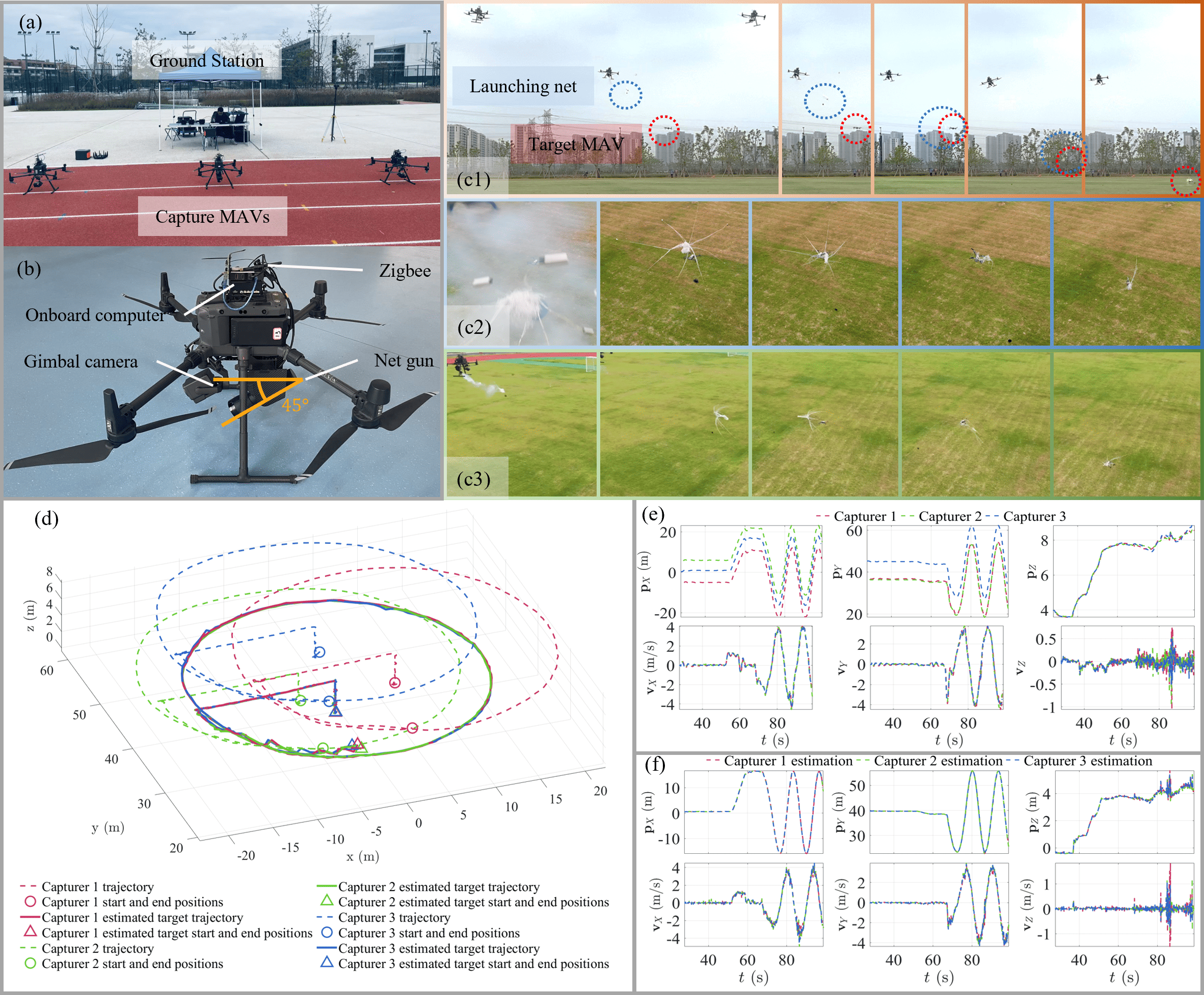}
    \caption{The outdoor experimental data of the cooperative MCM system pursuit and capture a target MAV with 4~m/s speed. 
    }
    \label{fig_exp}
\end{figure*}

\section{Simulation and Experiments Verification}

This section evaluates the performance of the cooperative MCM system in the \emph{Matlab} Simulation and numerous real-world experiments. The simulations evaluation comprises two key components: formation-based target pursuit with bearing-only measurements and net capture validation with a simplified net capture model. 

\subsection{Simulation results}

The simulation includes four pursuer MAVs and one target MAV. The target follows a set circular path with a radius of 10~m and a constant speed of 3~m/s. Each pursuer MAV acquires bearing measurements, which are subject to Gaussian noise (\(\epsilon_{\g} = 0.01 \, \text{rad}\)), and it also receives information including measurements and state estimates from one adjacent MAV at each time step. During the entire simulation, the pursuer MAVs are engaged in tasks such as gimbal tracking based on bearing measurements, state estimation, formation pursuing control, and autonomous capture determination.

The simulation results are depicted in Fig.~\ref{fig_sim_results}, including gimbal tracking, cooperative state estimation, tracking control, auto-capture evaluation, and flying net simulation. Figure~\ref{fig_sim_results}(a) displays the 3D flight paths of both the cooperative MCM system and the target MAV. Figure~\ref{fig_sim_results}(b) shows the gimbal tracking angle error between the camera's optical axis and the bearing vector to the target. Figure~\ref{fig_sim_results}(c) presents the target's estimated position and velocity errors by the four pursuer MAVs, respectively.  Fig.~\ref{fig_sim_results}(d) depicts the pursuer MAVs' attitude bias compared with static state and the capturable period determined by the real-time net model under the autonomous capture decisions of the four pursuer MAVs. The capturable zones of the cooperative MCM system effectively cover a significant portion of the target's flight path.

Furthermore, a flying net simulation with the complex net model is given to further illustrate the efficiency of the simplified capture zone as depicted in Fig.~\ref{fig_sim_results}(e). The simulation involves the fourth pursuer MAV, which initiates the net launch command at $t=20.50$~s. The flying net's states at 0.04~s, 0.14~s, and 0.24~s after launching are presented. It is evident that the target can be successfully captured with the simplified auto-launching determination.

\subsection{Experimental results}

The proposed cooperative MCM system demonstrated its effectiveness through numerous real-world trials. As shown in Fig.~\ref{fig_exp}(a), the cooperative MCM system consists of a ground station and three pursuer MAVs. Each pursuer MAV uses a DJI M300 as the platform, equipped with a DJI Manifold-2G onboard computer, an H20T gimbal camera, a Zigbee communicator, and a net-gun device, as depicted in Fig.~\ref{fig_exp}(b). The net gun is mounted on a \(45^\circ\) downward tilt.

\begin{table}[]
\centering 
\caption{The success rate of multiple experiments.}
\label{tab_success_rate}
\begin{tabular}{cccc}
\Xhline{1pt}
Success trials & Miss trials & Total trials & Success capture rate \\ \hline
11 & 6 & 17    & 64.7\%         \\
\Xhline{1pt}
\end{tabular}
\end{table}

The cooperative MCM system functions autonomously, executing all tasks onboard. During the experiment, the target MAV (DJI Mavic) traveled in a circular path with a 10~m radius at a speed of 4~m/s. Figure~\ref{fig_exp}(c1)-(c3) illustrate the capture process from the perspectives of the ground and MAVs 2 and 3, respectively. Figure~\ref{fig_exp}(d) shows the 3D flight paths of the pursuer MAVs and the estimated targets. Tracking positions and velocities of the pursuer MAVs along three axes are shown in Fig.~\ref{fig_exp}(e). Figure~\ref{fig_exp}(f) presents the estimated target's position and velocity results. These findings indicate the cooperative MCM system's proficiency in accurately pursuing and capturing targets using visual sensors. Multiple experiments were carried out, and the results are summarized in Table~\ref{tab_success_rate}, which reveals a capture success rate of $64.7\%$.

\section{Conclusion}

This paper presents a novel cooperative MAV capture (MCM) system that leverages standard optical sensors and operates without reliance on predefined unique features. By adopting a cooperative approach, the system effectively integrates target's state estimation with environmental tracking, enhancing its adaptability across various scenarios.  
Furthermore, a simplified flying net capture model is introduced, enabling real-time capture determination while significantly improving the system’s robustness and reliability in intercepting moving targets. The proposed cooperative MCM system has been extensively validated through both simulations and real-world experiments, demonstrating its effectiveness and practical applicability in dynamic operational environments.

\bibliographystyle{IEEEtran}
\bibliography{zcl_references}

\end{document}